\documentclass[letterpaper, 10 pt, conference]{ieeeconf}  % Comment this line out if you need a4paper
\usepackage{graphicx}
\usepackage{amsmath}
\usepackage{amssymb}
\usepackage{placeins}
\usepackage{subcaption} 
\usepackage{caption}
\usepackage{xcolor}
\usepackage{hyperref}
\hypersetup{
    colorlinks = true,
    linkcolor = blue,
    urlcolor = blue,
}

\usepackage{tikz}
\usetikzlibrary{shapes,arrows.meta,positioning,shadows.blur,calc}

\usepackage{booktabs}
\usepackage{cite}
\let\labelindent\relax
\usepackage{enumitem}
\usepackage{multirow}
\usepackage{tabularx}
\usepackage{algorithm}
\usepackage{algorithmicx}
\usepackage{algpseudocode}
\usepackage{url}
\usepackage{siunitx}

\newcommand{\nop}[1]{} 

\IEEEoverridecommandlockouts                              % This command is only needed if 
\title{\LARGE \bf
Joint Pedestrian and Vehicle Traffic Optimization in Urban Environments using Reinforcement Learning

}
\author{Bibek Poudel$^{1}$, Xuan Wang$^{2}$, Weizi Li$^{1}$, Lei Zhu$^{3}$, and Kevin Heaslip$^{4}$ % <-this % stops a space
\thanks{$^{1}$Bibek Poudel and Weizi Li are with Min H. Kao Department of Electrical Engineering and Computer Science at University of Tennessee, Knoxville, TN, USA {\tt\small bpoudel3@vols.utk.edu, weizili@utk.edu}}%
\thanks{$^{2}$Xuan Wang is with Department of Electrical and Computer Engineering at  George Mason University, Fairfax, VA, USA {\tt\small xwang64@gmu.edu}}%
\thanks{$^{3}$Lei Zhu is with Department of Industrial and Systems Engineering at University of North Carolina at Charlotte, Charlotte, NC, USA {\tt\small lei.zhu@charlotte.edu}}
\thanks{$^{4}$Kevin Heaslip is with Department of Civil and Environmental Engineering at University of Tennessee, Knoxville, TN, USA {\tt\small kheaslip@utk.edu}}
}

\begin{document}

\maketitle
\thispagestyle{empty}
\pagestyle{empty}

%Current approaches to traffic congestion mitigation with autonomous vehicles often overlook the complex dynamics of human-driven traffic. adress Sim-2-real gap.

%%%%%%%%%%%%%%%%%%%%%%%%%%%%%%%%%%%%%%%%%%%%%%%%%%%%%%%%%%%%%%%%%%%%%%%%%%%%%%%%
% \begin{abstract}

% Reinforcement learning (RL) holds significant promise for adaptive traffic signal control, yet existing RL-based methods primarily focus on vehicle-centric objectives, neglecting pedestrian efficiency and safety. In this paper, we introduce a deep RL framework for adaptive corridor-level control of eight traffic signals in a real-world urban network, jointly optimizing pedestrian and vehicular efficiency. Our unified policy, trained using real-world pedestrian and vehicle demand data, significantly reduces average wait times per pedestrian (up to $\mathbf{67\%}$) and per vehicle (up to $\mathbf{52\%}$), as well as total accumulated pedestrian and vehicle wait times (up to $\mathbf{67\%}$ and $\mathbf{53\%}$, respectively), compared to traditional fixed-time signals. Our results demonstrate the effectiveness of RL in balancing both pedestrian and vehicular needs within realistic urban traffic scenarios while maintaining strong generalization capabilities across varying traffic demands.

% \end{abstract}

\begin{abstract}

Reinforcement learning (RL) holds significant promise for adaptive traffic signal control. While existing RL-based methods demonstrate effectiveness in reducing vehicular congestion, their predominant focus on vehicle-centric optimization leaves pedestrian mobility needs and safety challenges unaddressed. In this paper, we present a deep RL framework for adaptive control of eight traffic signals along a real-world urban corridor, jointly optimizing both pedestrian and vehicular efficiency. Our single-agent policy is trained using real-world pedestrian and vehicle demand data derived from Wi-Fi logs and video analysis. The results demonstrate significant performance improvements over traditional fixed-time signals, reducing average wait times per pedestrian and per vehicle by up to $\mathbf{67\%}$ and $\mathbf{52\%}$ respectively, while simultaneously decreasing total wait times for both groups by up to $\mathbf{67\%}$ and $\mathbf{53\%}$. Additionally, our results demonstrate generalization capabilities across varying traffic demands, including conditions entirely unseen during training, validating RL's potential for developing transportation systems that serve all road users.

% Reinforcement learning (RL) holds significant promise for adaptive traffic signal control. However, existing RL-based methods primarily focus on vehicle-centric objectives, neglecting pedestrian mobility needs. In this paper, we introduce a deep RL framework for adaptive corridor-level control of eight traffic signals in a real-world urban network, jointly optimizing pedestrian and vehicular efficiency. Our policy is trained using real-world pedestrian and vehicle demand data derived from Wi-Fi logs and video analysis. This policy significantly reduces average wait times per pedestrian (up to $\mathbf{67\%}$) and per vehicle (up to $\mathbf{52\%}$), as well as total accumulated pedestrian and vehicle wait times (up to $\mathbf{67\%}$ and $\mathbf{53\%}$, respectively), compared to traditional fixed-time signals. Further, our results demonstrate generalization capabilities across varying traffic demands, including conditions entirely unseen during training.

\end{abstract}

%%%%%%%%%%%%%%%%%%%%%%%%%%%%%%%%%%%%%%%%%%%%%%%%%%%%%%%%%%%%%%%%%%%%%%%%%%%%%%%%
\section{Introduction}
\label{sec:intro}

\begin{figure*}[h!]
    \centering
    \includegraphics[width=\linewidth]{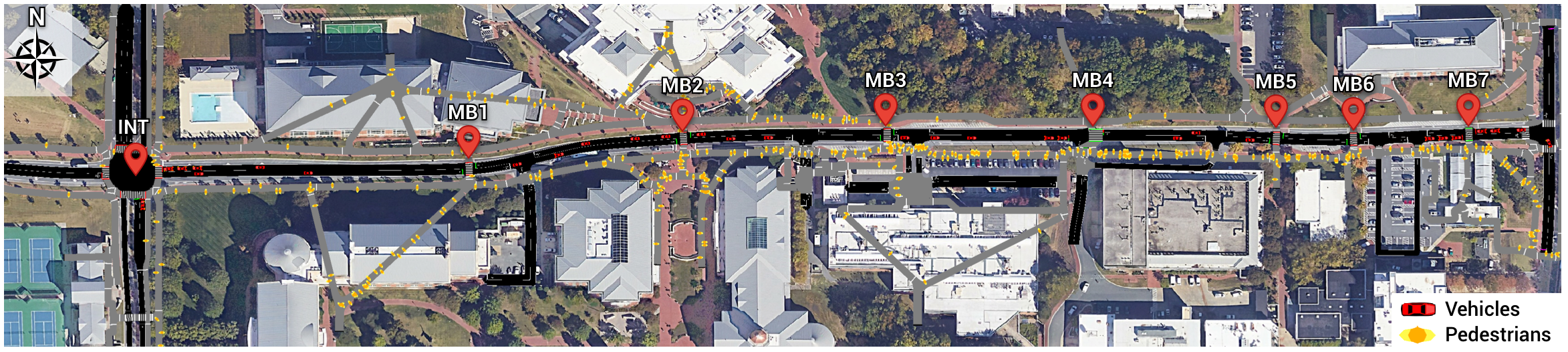} 
    \vspace{-1.5em}
    \caption{\small{The Craver Road corridor with an intersection (INT) that contains one primary traffic signal and four signalized crosswalks, along with seven mid-block signalized crosswalks (MB$1$–MB$7$) that control pedestrian–vehicle interactions.}}
    \label{fig:craver}
    \vspace{-1.3em}
\end{figure*}

Traffic congestion has become a silent tax on modern civilization. Each year, drivers in U.S. cities waste an average of $54$ hours stuck in traffic~\cite{TxDOT2023}, costing over $160$ billion in lost productivity~\cite{DOT2022}. With urban cores in metropolitan areas experiencing an increase of traffic inflow, up to $25\%$ year-over-year~\cite{INRIX2024}, congestion is set to worsen as urbanization continues. 
To address this challenge, traffic control systems have evolved from traditional handcrafted rules and actuated systems to Adaptive Traffic Signal Control (ATSC). Driven by cost-effectiveness~\cite{zhao2012overview, wang2024traffic}, increasing availability of traffic data~\cite{zhang2011data}, and advances in optimization techniques~\cite{xiao2021leveraging}, ATSC has become a central focus of intelligent transportation systems research~\cite{wei2021recent}. While ATSC has reduced congestion and improved vehicular flow, its evolution has largely overlooked a critical stakeholder: pedestrians.

Currently, pedestrian fatalities in the U.S. have reached their highest level in $41$ years, averaging about $21$ deaths per day~\cite{GHSA2023, NHTSA2023}. Urban areas are particularly affected, with $84$\% of these fatalities occurring in cities and $76$\% taking place at non-intersection locations such as mid-block crossings and commercial strips~\cite{NSC2023}. A key factor is the prevalence of unsignalized mid-block crosswalks, where pedestrians lack clear guidance and drivers may not yield, increasing accident risks~\cite{zegeer2001safety}. Implementing signalized crosswalks at these locations, with dedicated pedestrian phases and clear right-of-way signals, has proven effective in reducing vehicle-pedestrian conflicts and enhancing rule compliance~\cite{fitzpatrick2014characteristics}. The inclusion of pedestrian-friendly features in ATSC frameworks ensures that traffic control systems not only alleviate congestion but also proactively protect vulnerable road users.

As urban traffic control grows increasingly complex, ATSC strategies are leveraging real-time sensor data and computational methods. Reinforcement learning (RL) has emerged as a leading approach due to its ability to learn adaptive policies directly from data without relying on traffic models or handcrafted rules. 
Recent RL-based methods have primarily focused on improving vehicular throughput, neglecting pedestrian efficiency or modeling pedestrian behavior in overly simplified scenarios~\cite{zhang2018traffic, zheng2019learning, haydari2020deep}. This vehicle-centric approach fails to capture the complex interactions between vehicles and pedestrians, especially in urban environments characterized by high pedestrian volumes, frequent mid-block crossings, and diverse mobility demands. Developing control policies that simultaneously optimize pedestrian and vehicular efficiency in realistic urban settings remains an open challenge. 

We introduce an RL framework for corridor-level control of eight traffic signals, jointly optimizing for vehicular and pedestrian efficiency. Our contributions are

\begin{itemize}
    \item we develop a single policy that effectively manages high traffic volumes (up to $6,000$ pedestrians/hr and $558$ vehicles/hr) based on real-world demand data derived from Wi-Fi logs and video analysis;
    \item our results demonstrate substantial performance improvements over traditional fixed-time signals, reducing average wait time per vehicle by up to $52\%$ and average wait time per pedestrian by up to $67\%$; 
    \item we provide insights into the behaviors learned by our policy, including its ability to coordinate multiple independent signals to create a ``green wave'' effect and its responsiveness to real-time traffic, as demonstrated by its adaptive phase switching. 
\end{itemize}

\noindent The code, data, and videos are available in our GitHub: \href{https://github.com/poudel-bibek/Urban-Control}{\texttt{github.com/poudel-bibek/Urban-Control}}.

\section{Related Work}
\label{sec:related}

Conventional adaptive traffic control systems have relied on model-based~\cite{hunt1982scoot} or rule-based~\cite{sims1980sydney} approaches. Despite their widespread deployment and clear advantages over fixed-time control, these systems struggle to capture the inherent complexity and stochasticity of urban traffic. Common challenges include unpredictable variations in traffic volume and queue propagation through multiple intersections~\cite{papageorgiou2003review}. In response to these limitations, computational techniques that learn adaptive policies without relying on explicit rules or models, particularly machine learning and deep reinforcement learning (DRL), offer promising alternatives for traffic signal control~\cite{srinivasan2006neural, genders2016using, chu2019multi, wei2018intellilight, bie2024multi, genser2024time}. In single-agent settings, DRL has been applied to optimize intersection signal timing by selecting appropriate phase~\cite{mousavi2017traffic} or its duration~\cite{li2016traffic} based on observed traffic conditions. While multi-agent scenarios explore approaches where agents either collaborate to coordinate signals across a network~\cite{van2016coordinated} or compete to prioritize movement at a traffic signal with higher demand~\cite{zheng2019learning}. These methods demonstrate improved performance over fixed-time and conventional approaches~\cite{mannion2016experimental}, achieving, for instance, reductions in average travel time and queue lengths~\cite{liang2018deep}. Yet, it is worth noting that the majority of existing studies remain primarily focused on optimizing vehicle-centric outcomes such as reducing queue, decreasing stop frequency, and increasing throughput~\cite{haydari2020deep}.

% oroojlooy2020attendlight, chen2020toward, wiering2000multi

However, real-world urban environments include both vehicles and pedestrians. Integrating pedestrian dynamics into ATSC systems introduces several complex challenges such as ensuring pedestrian safety at crosswalks~\cite{zhang2018traffic, zhang2019pedestrian} and balancing the needs of both vehicles and pedestrians to prevent excessive delays for one group while prioritizing the other~\cite{han2022deep}. Recognizing these complexities, recent studies have extended RL-based traffic signal control to incorporate pedestrians. Several approaches have emerged in this direction: some introduce pedestrian-specific phases, i.e., eliminate vehicle-pedestrian interactions for improved safety~\cite{ma2014optimization}, while others incorporate pedestrian-centric performance metrics directly into the reward function~\cite{wu2020multi}. However, these studies typically rely on synthetic demand data or remain constrained to ideal road networks~\cite{wei2021recent}. This leaves pedestrian-inclusive ATSC strategies relatively underexplored, particularly for complex urban corridors with multiple signalized crosswalks.

Our work addresses these limitations by proposing a DRL framework that jointly optimizes vehicular and pedestrian waiting times in a real-world corridor-level setting. Unlike earlier methods that either focused solely on vehicles or used synthetic demand on ideal networks, our approach employs a real-world urban network of eight traffic signals with real-world demand data for both vehicles and pedestrians. The work most similar to ours uses a real-world corridor and vehicle demand data but differs by adopting a multi-agent framework and relying on synthetic pedestrian demand~\cite{jairam2024pedestrian}. Our approach demonstrates that a single-agent policy can effectively control an entire urban corridor while balancing the needs of both vehicles and pedestrians.

\section{Methodology}
\label{sec:methodology}
 
\begin{figure*}[t!]
    \centering
    \includegraphics[width=\linewidth]{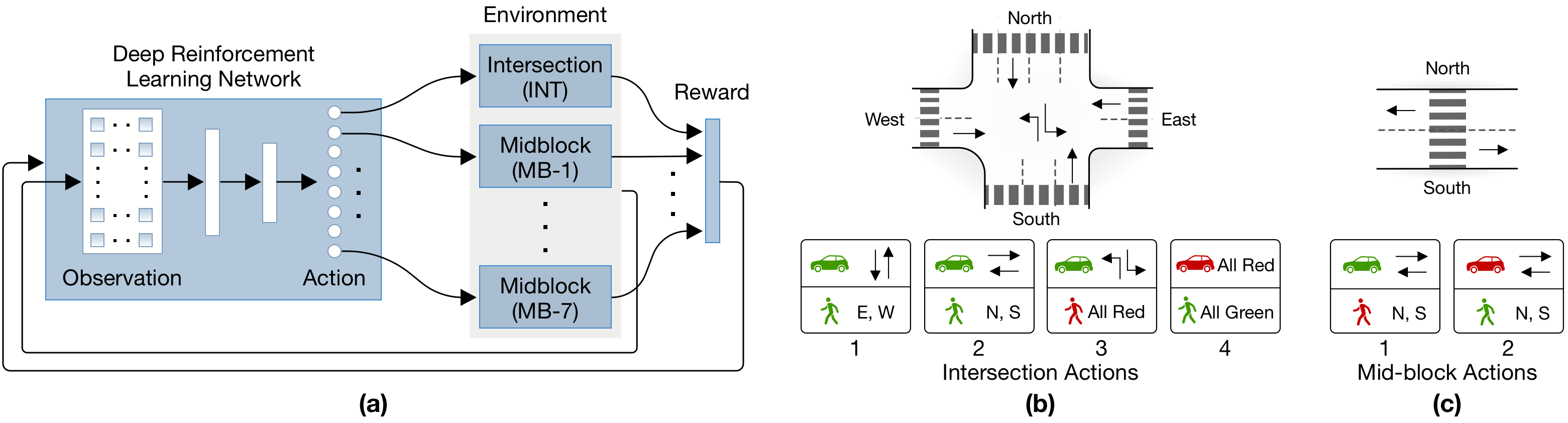} 
    \vspace{-1.5em}
    \caption{\small{(a) Deep reinforcement learning framework for corridor-level traffic control. (b) Intersection signal configurations controlling vehicle and pedestrian movements, including dedicated left turns (choice $3$) and all-pedestrian phases (choice $4$). (c) Two mid-block signal configurations allowing either vehicle movement (choice $1$) or pedestrian crossing (choice $2$).}}
    \label{fig:rl_actions}
    \vspace{-1.3em}
\end{figure*}

\subsection{Real-World Network and Traffic Data}

Craver Road, shown in Figure~\ref{fig:craver}, is a $750$m corridor that serves as the primary arterial through the University of North Carolina, Charlotte's main campus. The campus covers $1.56$ square miles and includes $85$ buildings and approximately $34,000$ students, faculty, and staff~\cite{unccharlotte_aboutus}. To capture pedestrian demand data, we utilize Wi-Fi logs collected in September $2021$ by the university's IT department. The Wi-Fi network comprises $2,492$ access points (APs) distributed across $82$ buildings, with $88,409$ unique clients. The Wi-Fi log data captures communication events between Wi-Fi clients (e.g., smartphones and laptops) and APs. Each log entry indicates ``when'' (timestamp) and ``where'' (AP location) each client connected to the network. We processed this raw data using a generalized Wi-Fi processing framework based on a ``Point-Line-Plane'' hierarchical concept~\cite{yuan2025hierarchical} with the following filtering assumptions:
\begin{itemize}
    \item Client activities were aggregated at the building level; each client's AP sessions within the same building were merged to represent presence in that location.
    \item Clients identified as visitors or irregular commuters (detected for fewer than three days per month, constituting $11.81\%$ of the dataset) were excluded, as our analysis focuses on typical campus travel patterns.
    \item To address individuals carrying multiple devices, we used K-means clustering to classify and remove ``non-mobile'' devices based on the mean and variance of their stationary ratio relative to total daily activity time~\cite{zhang2023crowdtelescope}.
\end{itemize}

\noindent For vehicle demand data, we analyzed four video recordings (total \(21\) minutes) captured at different times at the Craver Road intersection (INT). The observed flow was converted to hourly rates, resulting in an average headway of \(18\) seconds between vehicles. We mapped both pedestrian and vehicle demand to SUMO~\cite{krajzewicz2002sumo} trip definitions, creating a network-wide real-world demand of \(2223\) pedestrians/hr and \(202\) vehicles/hr. Of these pedestrians, $990$ ($44$\%) have trips that cross the corridor. For trip generation, we defined origin-destination pairs using Traffic Analysis Zones (TAZs). Pedestrian origin-destination pairs were derived from Wi-Fi building visit data, and vehicle pairs from movement records. In each trip, start and end points were assigned to specific edges within TAZs, with departure times based on observed timestamps.

\subsection{Markov Decision Process}
We formulate the Adaptive Traffic Signal Control (ATSC) as a sequential decision-making problem modeled as a partially observable Markov Decision Process represented by the tuple \((S, A, T, R, \Omega, O, \gamma)\). Here, \(S\) denotes the set of environment states, \(A\) represents the set of possible actions, \(T: S \times A \times S \to [0, 1]\) is the probabilistic state transition function, and \(R: S \times A \to \mathbb{R}\) defines the reward function. Due to partial observability in real-world traffic systems, \(\Omega\) represents the set of observations, \(O: S \times A \times \Omega \to [0, 1]\) defines the observation probability function, and \(\gamma \in [0, 1)\) is the discount factor to balance immediate and future rewards.

\textbf{State:} Traffic representation significantly impacts ATSC performance~\cite{zhang2022expression}. Our observation fuses vehicle and pedestrian occupancy data in areas neighboring each traffic signal. Vehicles are detected within a $15$--$100$m vicinity while pedestrians are detected within $5$--$10$m. To capture both spatial and temporal dynamics, we stack occupancy information over the action duration (each action step encompasses \(N\) simulation timesteps). At action step \(t\), the observation \(o_t \in \Omega\) is formed by stacking \(N\) vectors from the previous action interval:
\begin{align*}
o_t &= \left[ v_1,\, v_2,\, \ldots,\, v_N \right], \quad \text{with} \\
v_k &= \Bigl[\phi_{t-1},\, \{v_{i,\cdot}(k)\}_{i=1}^{M},\, \{p_{i,\cdot}(k)\}_{i=1}^{M}\Bigr],
\end{align*}
for \(k = 1, 2, \ldots, N\), where:
\begin{itemize}
    \item \(\phi_{t-1}\) is the signal phase during the previous action,
    \item \(v_{i,\cdot}(k) = \bigl(v_{i,\mathrm{in}}(k),\, v_{i,\mathrm{inside}}(k),\, v_{i,\mathrm{out}}(k)\bigr)\) denotes the vehicle occupancy vector at traffic signal \(i\) at the \(k\)-th simulation timestep; here, \(v_{i,\mathrm{in}}(k)\) represents occupancy from lanes approaching the traffic signal (incoming), \(v_{i,\mathrm{inside}}(k)\) from lanes within the controlled area, and \(v_{i,\mathrm{out}}(k)\) from lanes exiting (outgoing),
    \item \(p_{i,\cdot}(k) = \bigl(p_{i,\mathrm{in}}(k),\, p_{i,\mathrm{out}}(k)\bigr)\) denotes the pedestrian occupancy vector at traffic signal \(i\) at the \(k\)-th simulation timestep; here, \(p_{i,\mathrm{in}}(k)\) represents occupancy from pedestrians approaching the crosswalk, and \(p_{i,\mathrm{out}}(k)\) from those leaving,
    \item \(M\) is the number of controlled traffic signals.
\end{itemize}
This spatio-temporal formulation provides a comprehensive snapshot of evolving traffic conditions.

\textbf{Action:} The agent's action is composed of two independent components—intersection and mid-block actions—both derived from a policy parameterized by \(\theta \in \mathbb{R}^d\), which takes as input the current observation \(o_t\) to produce logits that are split into distribution parameters for the two actions:
\begin{itemize}
    \item \emph{Intersection Action}: The agent selects one of four mutually exclusive phase configurations as shown in Figure~\ref{fig:rl_actions} (b), i.e., \(a_t^{int} \in \{1,2,3,4\}\), with \(j\) denoting the chosen configuration. This selection is drawn from a Categorical distribution:
    \[
    P(a_t^{int} = j \mid o_t, \theta) = \text{Categorical}\Bigl(j; \mathbf{p}^{int}(o_t, \theta)\Bigr),
    \]
    where \(\mathbf{p}^{int}(o_t, \theta) \in \Delta^{3}\) is the probability vector over the four phase configurations.
    
    \item \emph{Mid-Block Actions}: For each of the seven mid-block signals, the agent independently selects a binary action \(a_{t,i}^{mb} \in \{1,2\}\), each modeled as a Bernoulli distribution:
    \[
    P(a_{t,i}^{mb} = b \mid o_t, \theta) = \text{Bernoulli}\Bigl(b; \mu_i^{mb}(o_t, \theta)\Bigr),
    \]
    where \(\mu_i^{mb}(o_t, \theta)\) is the probability for signal \(i\) and \(b=1\) indicates permission for vehicle flow, i.e., Mid-block choice $1$ shown in Figure~\ref{fig:rl_actions} (c).
\end{itemize}

The overall action \(a_t = \text{concat}(a_t^{int}, a_{t,1}^{mb}, \ldots, a_{t,7}^{mb})\) is an \(8\)-component vector. Restricting the intersection action to pre-defined safe configurations inherently enforces safety (preventing conflicting pedestrian–vehicle greens) and reduces the exploration space for more sample-efficient learning. For additional safety, a \(4\)-timestep mandatory yellow phase is automatically introduced for all signals via an internal mechanism before switching from a green to a red signal. Note that while direct control over the yellow phase duration is not permitted, the duration of any other phase can be controlled by repeatedly selecting the same action. Each action lasts for $10$ simulation steps (\(N=10\)).

\textbf{Reward:} To simultaneously minimize wait times for vehicles and pedestrians, we propose the \emph{Exponentially Increasing Maximum Wait Aggregated Queue (EI-MWAQ)} reward function. This reward is designed to reflect real-world behavior, where very long wait times lead to disproportionately high driver and pedestrian frustration. It builds on the Maximum Wait Aggregated Queue (MWAQ)~\cite{koohy2022reward}, which uses the product of queue length and maximum waiting time to approximate worst-case delay, but introduces two modifications. First, penalties from all mid-block crossings are aggregated using an $L_2$-norm. Second, the final penalty values are obtained by normalizing these aggregate delays and applying an exponential function. This results in a penalty that grows rapidly as queue lengths and wait times increase. For the intersection, we compute:

\[
Q^{\mathrm{int}}_{\mathrm{veh}} = \frac{N^{\mathrm{int}}_{\mathrm{veh}} \cdot W^{\mathrm{int}}_{\mathrm{veh}}}{8|D|}, \quad 
Q^{\mathrm{int}}_{\mathrm{ped}} = \frac{N^{\mathrm{int}}_{\mathrm{ped}} \cdot W^{\mathrm{int}}_{\mathrm{ped}}}{10|D|},
\]
where \(N^{\mathrm{int}}_{\mathrm{veh}}\) and \(N^{\mathrm{int}}_{\mathrm{ped}}\) denote the counts of waiting vehicles and pedestrians at the intersection, \(W^{\mathrm{int}}_{\mathrm{veh}}\) and \(W^{\mathrm{int}}_{\mathrm{ped}}\) their respective maximum waiting times, and \(|D|\) the number of incoming directions. For each mid-block signal \(i\), we compute:
\begin{align*}
Q^{\mathrm{mb}}_{\mathrm{veh}}(i) &= \frac{N^{\mathrm{mb}}_{\mathrm{veh}}(i) \cdot W^{\mathrm{mb}}_{\mathrm{veh}}(i)}{8|D_{\mathrm{mb}}|}, \\
Q^{\mathrm{mb}}_{\mathrm{ped}}(i) &= \frac{N^{\mathrm{mb}}_{\mathrm{ped}}(i) \cdot W^{\mathrm{mb}}_{\mathrm{ped}}(i)}{10},
\end{align*}
where \(|D_{\mathrm{mb}}|\) is the number of incoming directions. These per-signal values are aggregated across all \(7\) mid-block signals using the \(L_2\) norm:
\[
Q^{\mathrm{mb}}_{\mathrm{veh}} = \left\|\left(Q^{\mathrm{mb}}_{\mathrm{veh}}(i)\right)_{i=1}^{7}\right\|_2, \quad 
Q^{\mathrm{mb}}_{\mathrm{ped}} = \left\|\left(Q^{\mathrm{mb}}_{\mathrm{ped}}(i)\right)_{i=1}^{7}\right\|_2.
\]
The amplified penalties are obtained by applying the exponential function:
\[
\begin{aligned}
R^{\mathrm{int}}_{\mathrm{veh}} &= \exp\left(Q^{\mathrm{int}}_{\mathrm{veh}}\right), \quad
R^{\mathrm{int}}_{\mathrm{ped}} = \exp\left(Q^{\mathrm{int}}_{\mathrm{ped}}\right), \\
R^{\mathrm{mb}}_{\mathrm{veh}} &= \exp\left(Q^{\mathrm{mb}}_{\mathrm{veh}}\right), \quad
R^{\mathrm{mb}}_{\mathrm{ped}} = \exp\left(Q^{\mathrm{mb}}_{\mathrm{ped}}\right).
\end{aligned}
\]
The final reward is given by:
\[
\boxed{ R = -\left(R^{\mathrm{int}}_{\mathrm{veh}} + R^{\mathrm{int}}_{\mathrm{ped}} + R^{\mathrm{mb}}_{\mathrm{veh}} + R^{\mathrm{mb}}_{\mathrm{ped}}\right) },
\]
which is clipped within the range \([-10^5,\,0]\) for numerical stability. Vehicles are considered waiting below a speed of \(0.2\,\mathrm{m/s}\) and pedestrians below a speed of \(0.5\,\mathrm{m/s}\), and the normalization constants \(8\) and \(10\) are empirically chosen. Additionally, both the state and reward statistics are updated at each action step using a Welford Normalizer~\cite{huang202237, rl_bag_of_tricks}.

\section{Experiments}
\label{sec:experiment}

\begin{table}[t!]
\begin{center}
  \small
  \setlength{\tabcolsep}{4pt}
  \begin{tabular}{llr}
    \toprule
    Category & Parameter & Value \\
    \midrule    
    \multirow{7}{*}{PPO} 
    & Learning Rate ($\alpha$) & $1{\times}10^{-4}$ \\
    & Discount Factor ($\gamma$) & $0.99$ \\
    & GAE Estimation ($\lambda$) & $0.95$ \\
    & Clip Parameter ($\epsilon$) & $0.2$ \\
    & Value Function Coeff. & $0.5$ \\
    & Update Frequency & $1024$ \\
    & K-epochs & $4$ \\
    \midrule
    \multirow{3}{*}{Policy} 
    & Architecture & MLP \\
    & Hidden Layers (Actor) & $[512, 256, 128, 64, 32]$ \\
    & Hidden Layers (Critic) & $[512, 256, 128, 64, 32]$ \\
    % & Activation & Tanh \\
    \midrule
    \multirow{10}{*}{Simulation} 
    & Crossing Width & $4$~m\\ 
    & Sidewalk Width & $4$~m\\ 
    & Time Step ($\Delta t$) & $1$ second \\
    & Action Duration & $10$ steps \\
    & Warmup Timesteps & $100$-$250$ \\
    & Episode Horizon & $600$ steps \\
    & Vehicle Control Model & IDM~\cite{treiber2013traffic}\\
    & Pedestrian Control Model & Stripping~\cite{erdmann2015modelling}\\
    & Vehicle Speed Limit & $50$~km/hr\\
    & Pedestrian Walking Speed & $2.78$~m/s\\
    \bottomrule
  \end{tabular}
\end{center}
\vspace{-6pt}
\caption{\small{Key parameters for PPO, policy, and simulation.}}
\vspace{-10pt}
\label{table:params}
\end{table}

\subsection{Setup}
We conduct all simulations using SUMO~\cite{krajzewicz2012recent}. The reinforcement learning policy is trained using Proximal Policy Optimization (PPO)~\cite{schulman2017proximal} with Generalized Advantage Estimation (GAE)~\cite{schulman2015high}. Training occurs across $24$ parallel actors over $6 \times 10^6$ simulation timesteps on an Intel Core i\(9-14900\)KF processor and an NVIDIA RTX A\(5000\) GPU. We implement a multi-layer perceptron policy architecture with separate networks for actor and critic. During training, each episode consists of a warmup period (randomly selected between $100$ and $250$ timesteps) during which all signals operate on fixed-time control, followed by a $600$-timestep episode horizon. To ensure robust policy learning, we randomly scale both pedestrian and vehicle demand between $1\times$ and $2.25\times$ the original demand for each episode. SUMO's dynamic routing behavior introduces additional variability by adapting vehicle and pedestrian routes based on current traffic conditions. A comprehensive list of simulation and training parameters is provided in Table~\ref{table:params}.

\begin{figure*}[t!]
    \centering
    \includegraphics[width=\linewidth]{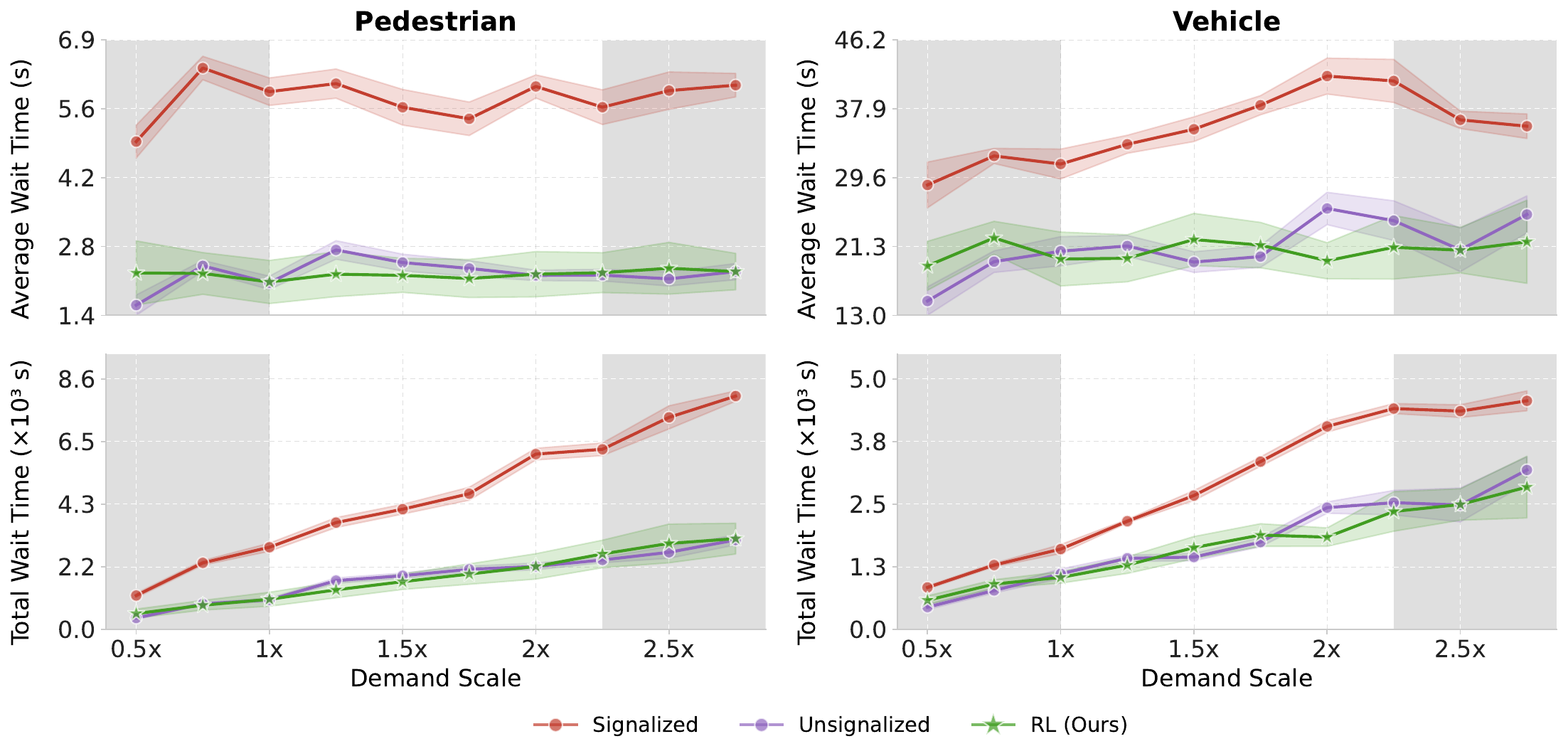} 
    \vspace{-1.5em}
    \caption{\small{Performance comparison between three traffic control approaches at various demands. The figure shows wait times for both pedestrians (left) and vehicles (right), measured as average wait time per pedestrian/vehicle (top) and total wait time for pedestrians/vehicles (bottom). Our RL agent consistently outperforms the fixed-time signal control (Signalized) across all demand levels, reducing average vehicle wait times by up to $52\%$ (from $42.6$ to $20.5$ seconds per vehicle at $2$× demand) while decreasing pedestrian wait times by up to $67\%$ (from $6.0$ to $2.0$ seconds per pedestrian at $2$× demand). Despite providing protected crossing phases through signalization, our RL approach achieves pedestrian wait times comparable to or slightly better than unsignalizing mid-block crosswalks (Unsignalized), while delivering approximately $20\%$ lower average vehicle wait times at higher demands. The gray-shaded areas to the left and right indicate demand levels unseen during training ($<1$× and $>2.25$×), where our approach generalizes effectively with consistent performance improvements across both low and high demands. All values are averaged across $10$ independent simulation runs with a total of $600$ runs.}}
    \label{fig:consolidated_results}
    \vspace{-1.3em}
\end{figure*}

\subsection{Benchmarks}
We evaluate our approach against two traffic control strategies that represent common real-world implementations.

\subsubsection{Unsignalized}
\begin{figure*}[t!]
    \centering
    \includegraphics[width=0.96\linewidth]{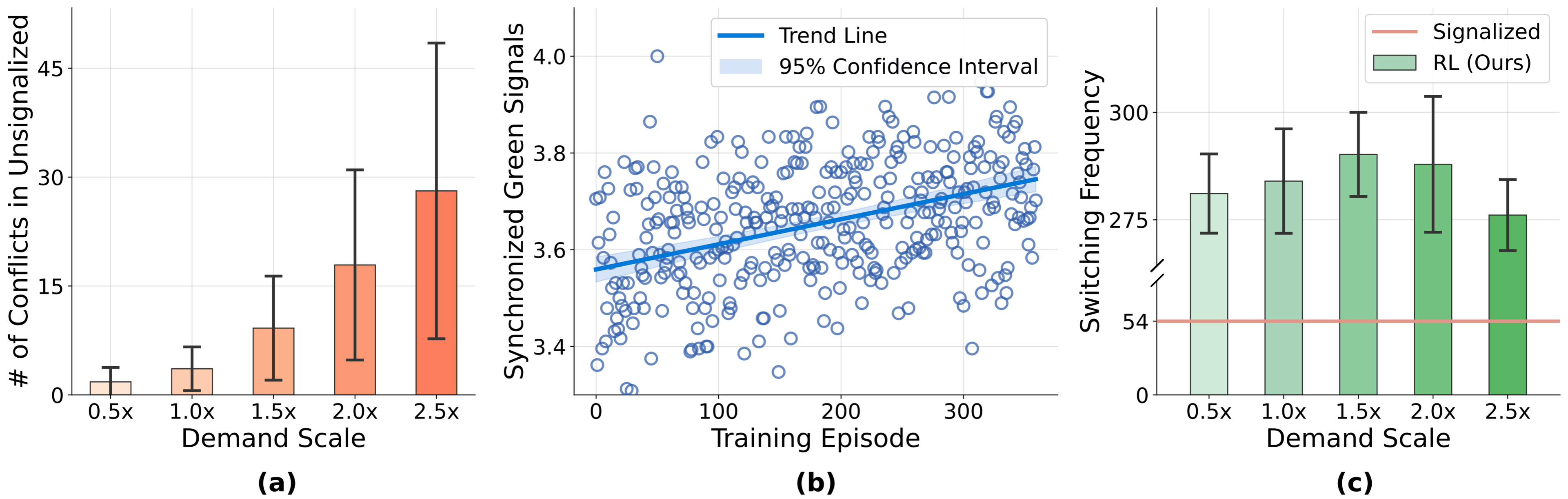} 
    \vspace{-2pt}
    \caption{\small{(a) Vehicle-pedestrian right-of-way conflicts in unsignalized mid-block crosswalks increase substantially with traffic demand, from \(1.8\) conflicts at \(0.5\)x demand to \(28.1\) at \(2.5\)x demand. The error bars show standard deviation values rising at higher demand levels, indicating safety outcomes become both worse and more unpredictable as traffic volumes increase. (b) Emergence of traffic signal coordination behavior during training. The average number of mid-block traffic lights simultaneously set to green for vehicular flow shows an upward trend (from approximately $3.55$ to $3.75$) indicating that the RL agent gradually learns to coordinate multiple signals. (c) Total signal phase switches after the warmup period across all signals in the network. RL agent demonstrates adaptive switching, with more than fivefold increase in switching frequency compared to fixed-time signalized control. Plots (a) and (c) data are averaged over \(10\) runs.}}
    \label{fig:insights}
    \vspace{-1.3em}
\end{figure*}

In this benchmark, we implement all mid-block locations (MB$1$-MB$7$) as unsignalized crosswalks, closely matching the current real-world setup of the corridor. At these unsignalized crosswalks, pedestrians have the right-of-way as specified by the Uniform Vehicle Code~\cite{UVC2000}. This right-of-way behavior is implemented in our simulation using SUMO's pedestrian interaction model~\cite{sumo_pedestrians}, where vehicles must yield to pedestrians in two specific scenarios:
\begin{itemize}
    \item Vehicles and pedestrians share the same road
    \item Vehicles pass through designated pedestrian crossings
\end{itemize}
The unsignalized approach represents a pedestrian-prioritized baseline that minimizes pedestrian delay at mid-block locations but may increase vehicle delay and conflicts (right-of-way negotiations) between vehicles and pedestrians. As shown in Figure~\ref{fig:insights}(a), the conflicts become increasingly critical at higher traffic volumes, where unsignalized crosswalks experience up to \(28.1\) conflicts on average. The intersection (INT) remains signalized.

\subsubsection{Signalized}  
This benchmark implements fixed-time control for both the intersection (INT) and mid-block (MB$1$-MB$7$) crosswalks. The signal timings are based on real-world observations and standard traffic engineering practices:

\textbf{INT:} Operates on a $5$-phase cycle with $90$-second green periods alternating between N-S and E-W through vehicle movements. Each direction change includes a $4$-second yellow and $2$-second all-red transition period. Consistent with the real-world implementation, the signal timing does not include dedicated left-turn phases. Complementary pedestrian crossings activate simultaneously with their corresponding vehicle phases (i.e., when N-S vehicle movement is green, the E-W pedestrian crosswalks are also green, and vice versa). We derived these timings through manual observation of video footage captured at different times of day.

\textbf{MB$\mathbf{1}$-MB$\mathbf{7}$:} Operate on a $62$-second cycle with phases set according to guidelines from FHWA's Manual on Uniform Traffic Control Devices (MUTCD)~\cite{traffic2023manual} and Traffic Signal Timing Manual (TSTM)~\cite{koonce2008traffic}:  
\begin{itemize}  
    \item Pedestrian Phase (MUTCD 4I$.06$): $16$ seconds consisting of a $7$-second minimum interval followed by a $9$-second clearance interval calculated as:
    \begin{align*}  
        \text{Clearance time} = \frac{\text{Crosswalk length}}{\text{Walking speed}} = \frac{32 \text{ ft}}{3.5 \text{ ft/s}} \approx 9 \text{ s}  
    \end{align*}  
    \item Vehicle Phase (TSTM $6.6.3$, MUTCD 4F$.17$): $46$ seconds consisting of $40$-second green time ($64\%$ of the split distribution), a $4$-second yellow change interval, and a $2$-second red clearance interval.  
\end{itemize}  

The signalized approach represents a fully-controlled safety-oriented baseline with dedicated signal phases ensuring rule compliance and eliminating vehicle-pedestrian right-of-way conflicts. Our RL approach also uses the fully-controlled setup but replaces the fixed signal timing cycles with adaptive timings controlled by the policy.

\subsection{Results}
Our reinforcement learning (RL) based approach effectively resolves the safety-efficiency trade-off in traffic control by providing the safety benefits of signalized crossings while achieving wait times comparable to or better than unsignalized crossings. As shown in Figure~\ref{fig:consolidated_results}, the approach consistently outperforms \textit{Signalized} across all demand levels, reducing average vehicle wait times by up to $52\%$ and pedestrian wait times by up to $67\%$. The approach also demonstrates strong generalization capabilities, maintaining these performance advantages at both low ($0.5$x, $0.75$x) and high ($2.5$x, $2.75$x) demands that were unseen during training.

\textbf{Pedestrian Wait Times.} 
At $1$x demand, our approach achieves an average wait time per pedestrian of $2.1$ seconds compared to $5.9$ seconds for \textit{Signalized} and $2.1$ seconds for \textit{Unsignalized}. This represents a $65\%$ reduction compared to \textit{Signalized} while matching \textit{Unsignalized}. As demand increases, our approach maintains its efficiency: at $2$x demand, $2.0$ seconds versus $6.0$ seconds for \textit{Signalized} ($67\%$ reduction) and $2.2$ seconds for \textit{Unsignalized}. The total pedestrian wait time shows a similar pattern (up to $67\%$ reduction), with our approach consistently outperforming both \textit{Signalized} and \textit{Unsignalized} across the demands. Even as pedestrian demand increases, our agent consistently maintains efficient crossing opportunities without compromising safety (no pedestrian-vehicle conflicts).

\textbf{Vehicle Wait Times.} 
At $1$x demand, our approach achieves an average wait time per vehicle of $20.8$ seconds compared to $31.3$ seconds for \textit{Signalized} and $20.7$ seconds for \textit{Unsignalized}. This represents a $34\%$ reduction compared to \textit{Signalized} while maintaining comparable performance to \textit{Unsignalized}. This pattern continues at higher demands: at $2$x demand, $20.5$ seconds versus $42.6$ seconds for \textit{Signalized} ($52\%$ reduction) and $25.9$ seconds for \textit{Unsignalized}. The total vehicle wait time at $2$x demand shows similar improvements: $1.16$ hours for \textit{Signalized}, $0.55$ hours for our approach, and $0.56$ hours for \textit{Unsignalized}.

\textbf{Generalization to Unseen Demands.} 
As shown in the shaded (gray) regions of Figure~\ref{fig:consolidated_results}, our approach maintains consistent performance advantages for both pedestrians and vehicles at both low demand (up to $0.5$x) and high demand (up to $2.75$x), reducing wait times by up to $67$\% for pedestrians and up to $39$\% for vehicles.

\noindent To understand the observed wait time reductions, we examine the internal behaviors of our policy and identify two emergent phenomena that explain its advantage over baselines:

\textbf{Signal Coordination.} 
Our policy autonomously learns to coordinate mid-block crosswalks, similar to the ``green wave'' effect observed in other RL traffic systems~\cite{wei2019deep, liu2023deep}. The action space models each mid-block signal independently using a Bernoulli distribution (as described in Section~\ref{sec:methodology}), with no built-in mechanism for coordination. Yet as shown in Figure~\ref{fig:insights}(b), the RL agent learns to increase the number of mid-block traffic signals simultaneously set to vehicle green phase. The trend line shows this coordination increasing from approximately $3.55$ to $3.75$ synchronized green signals over the course of training—a significant shift when considering each data point represents an average over $1440$ actions taken across $24$ parallel actors. The coordination of signal timing reduces the number of stops and enables more efficient vehicular flow through the corridor.

\textbf{Adaptive Switching Frequency.}
As shown in Figure~\ref{fig:insights}(c), our policy exhibits more than fivefold increase in switching frequency with an average of $284$ switches compared to $54$ in signalized control. Additionally, the policy demonstrates adaptability across different demand scales, with standard deviations ranging from \(9.23\) (at $2.5$x demand) to $16.42$ (at $2.0$x demand). Despite the policy not being explicitly incentivized to switch more (or less) often in the reward, it learned that demand-responsive and generally more frequent switching enables the system to minimize waiting times.

\section{Conclusion and Future Work}
\label{sec:conclusion}

In this work, we introduced a deep reinforcement learning framework for corridor-level adaptive traffic signal control that jointly optimizes for pedestrian and vehicular efficiency. We applied this framework to a real-world urban network using real-world pedestrian and vehicle demand data. Our trained policy reduces wait times significantly for both pedestrians (up to $67\%$) and vehicles (up to $52\%$) compared to traditional fixed-time signals, while generalizing to both lower and higher traffic volumes not seen during training. 

% Limitations.
While our approach demonstrates significant improvements, several limitations exist. First, under high demand, we observe vehicle queue propagation upstream when signals are closely spaced (a back-spill effect). This is likely a result of our modeling assumption, i.e., we assume signals operate independently. Second, the high switching frequency exhibited by our policy, while reducing wait times, breaks traffic flow continuity and may increase vehicle energy consumption because of more frequent stop-and-go events. Third, our evaluation was limited to a university campus setting with its specific pedestrian and vehicle patterns; generalization to heterogeneous traffic conditions, different corridor geometries including irregular intersections, and diverse urban environments remains to be validated. Fourth, while our baselines include common real-world approaches (fixed-time control and unsignalized crosswalks), direct comparison with other RL-based controllers proved challenging: existing RL methods either exclude pedestrian considerations entirely or require separate agents at each intersection with distributed training infrastructure and communication protocols, whereas our approach uses a single policy controlling all eight signals through centralized training.

Future work could address these limitations by explicitly modeling correlations between adjacent signals, incorporating traffic flow continuity or energy consumption metrics into the reward function, and evaluating performance across diverse urban settings with varying pedestrian-vehicle ratios and geometric configurations. Additional future directions include mixed-traffic control~\cite{wang2024learning, villarreal2024mixed}, adversarial robustness evaluation~\cite{poudel2021black}, and real-world traffic disruption analysis~\cite{poudel2024endurl, poudel2024carl}.

\section*{Acknowledgments}
This research is supported by NSF IIS-$2153426$ and ECCS-$2332210$. The authors also thank NVIDIA and the Tickle College of Engineering at University of Tennessee, Knoxville for their support.

\bibliographystyle{unsrt}
\bibliography{references}

% \newpage
% \thispagestyle{empty}
% \mbox{}

% \input{sections/appendix}

% \section*{ACKNOWLEDGMENT}
% The preferred spelling of the word ÒacknowledgmentÓ in America is without an ÒeÓ after the ÒgÓ. Avoid the stilted expression, ÒOne of us (R. B. G.) thanks . . .Ó  Instead, try ÒR. B. G. thanksÓ. Put sponsor acknowledgments in the unnumbered footnote on the first page.

\end{document}